\pdfoutput=1

\documentclass[11pt]{article}

\usepackage[final]{acl}

\usepackage{times}
\usepackage{latexsym}
\usepackage{booktabs} 
\usepackage{amsmath,amssymb,amsthm}  
\usepackage{enumitem}                
\usepackage{comment}
\usepackage{xcolor}
\usepackage{soul}
\usepackage{ulem}

\usepackage[T1]{fontenc}

\usepackage[utf8]{inputenc}
\usepackage{comment}
\usepackage{microtype}

\usepackage{inconsolata}

\usepackage{graphicx}

\title{BASIR: Budget-Assisted Sectoral Impact Ranking - A Dataset for Sector Identification and Performance Prediction Using Language Models}

\author{Sohom Ghosh \\
  Jadavpur University \\
  Kolkata, India \\
  \texttt{sohom1ghosh@gmail.com} \\\And
  Sudip Kumar Naskar \\
  Jadavpur University \\
  Kolkata, India \\
  \texttt{sudip.naskar@gmail.com} \\}

\begin{document}
\maketitle
\begin{abstract}
Government fiscal policies, particularly annual union budgets, exert significant influence on financial markets. However, real-time analysis of budgetary impacts on sector-specific equity performance remains methodologically challenging and largely unexplored. This study proposes a framework to systematically identify and rank sectors poised to benefit from India’s Union Budget announcements. The framework addresses two core tasks: (1) multi-label classification of excerpts from budget transcripts into 81 predefined economic sectors, and (2) performance ranking of these sectors. Leveraging a comprehensive corpus of Indian Union Budget transcripts from 1947 to 2025, we introduce \textbf{BASIR} (\textbf{B}udget-\textbf{A}ssisted \textbf{S}ectoral \textbf{I}mpact \textbf{R}anking), an annotated dataset mapping excerpts from budgetary transcripts to sectoral impacts. Our architecture incorporates fine-tuned embeddings for sector identification, coupled with language models that rank sectors based on their predicted performances. Our results demonstrate 0.605 F1-score in sector classification, and 0.997 NDCG score in predicting ranks of sectors based on post-budget performances. The methodology enables investors and policymakers to quantify fiscal policy impacts through structured, data-driven insights, addressing critical gaps in manual analysis. The annotated dataset has been released under CC-BY-NC-SA-4.0 license to advance computational economics research.
\end{abstract}

\section{Introduction}
\begin{figure}[t]
  \includegraphics[width=\columnwidth]{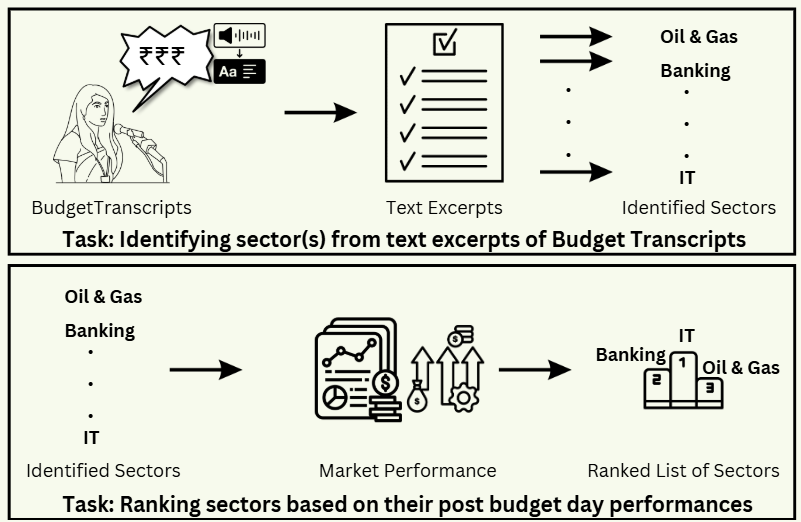}
  \caption{Identifying and Ranking sectors from transcripts of Indian Union Budgets
    }
  \label{fig:budget-intro}
\end{figure}

In emerging economies, government budget plans greatly influence financial markets. \footnote{\href{https://www.ndtv.com/india-news/explained-how-union-budget-influences-stock-market-7517036}{https://www.ndtv.com/india-news/explained-how-union-budget-influences-stock-market-7517036} (accessed on 16\textsuperscript{th} March, 2025)} In India, the Union Budget’s sector-specific allocations and tax reforms directly influence capital flows, with historical data showing volatility spikes in sectoral indices during budget weeks. \footnote{\url{https://cleartax.in/s/budget-day-market-movement-history-in-india} (accessed on 16\textsuperscript{th} March, 2025)} Investors systematically scrutinize budgetary provisions to predict market trajectories. \footnote{\href{https://economictimes.com/markets/stocks/news/consumption-over-capex-how-the-budget-impacts-stock-market-investors/articleshow/117853360.cms}{https://economictimes.com/markets/stocks/news/consumption-over-capex-how-the-budget-impacts-stock-market-investors/articleshow/117853360.cms} (accessed on 16\textsuperscript{th} March, 2025)} Existing works demonstrates significant correlations between budgetary measures and sectoral performance, particularly in consumption-driven industries. \footnote{\href{https://economictimes.com/markets/stocks/news/budget-2025-impact-on-stock-market-which-sectors-stand-to-benefit-or-lose/sector-trends/slideshow/117834567.cms}{https://economictimes.com/markets/stocks/news/budget-2025-impact-on-stock-market-which-sectors-stand-to-benefit-or-lose/sector-trends/slideshow/117834567.cms} (accessed on 16\textsuperscript{th} March, 2025)} However, current analysis methodologies remain predominantly manual, which is labour-intensive, time consuimg, prone to cognitive biases, and may be based on speculation. This study addresses these gaps using  a novel computational framework combining transformer-based language models with sectoral performance ranking. We present the tasks in Figure \ref{fig:budget-intro}.\\
Our contributions include:\\
$\bullet$ \textbf{BASIR} (\textbf{B}udget-\textbf{A}ssisted \textbf{S}ectoral \textbf{I}mpact \textbf{R}anking) – The first annotated dataset spanning Indian Union Budgets from the year 1947 to 2025, featuring 1,600+ texts from budget transcripts with corresponding sectors labeled. Furthermore,  we present 400+ texts with thier corresponding sectoral performance post the day of budget announcement.\\
$\bullet$ A framework for identifying sectors from budget transcripts and ranking them based on predicted performance.\\
$\bullet$ Empirical assessment of advanced Large Language Models' capabilities in predicting performance of sectors based on text excerpts from budget transcripts.


\section{Related Work}
The annual Indian Union Budget functions as a crucial instrument for economic policymaking, exerting a direct impact on sectoral growth trajectories and investor sentiment within equity markets \cite{panwar2019impact}. 
Research using event study methodology has demonstrated that Cumulative Average Abnormal Returns (CAARs) are significant around budget announcements, indicating that these events contain valuable information for market participants \cite{kharuristock} \cite{manjunatha2023effects}.  Our research investigates the empirical evidence of budget-induced stock market reactions across major sectors, with a focus on the transformative role of NLP in this.

Studies \cite{martin2024analyzing} \cite{joshi2018impact} reveal pronounced sector-specific volatility patterns post-budget announcements, with healthcare, banking, and Information Technology (IT) sectors demonstrating heightened sensitivity to tax reforms and capital allocation decisions. 
\cite{mansurali2022social} worked on analysisng sentiments of tweets relating to Budget 2020. NLP has emerged as a transformative tool in decoding the impacts of fiscal policy on stock markets.
Sentiment analysis, a subfield of NLP, is particularly useful in assessing market sentiment and generating trading signals based on prevailing trends \cite{saxena2021stock}. 
For instance, advanced NLP models like BERTopic \cite{grootendorst2022bertopic} and RoBERTa \cite{Liu2019RoBERTaAR} have been employed to analyze the sentiment of the Reserve Bank of India's monetary policy communications, revealing how different economic topics influence market reactions \cite{kumar2024words}. 

Most previous studies have focused on post-hoc analyses using historical data, typically conducted after market hours. Our work, however, introduces a predictive approach, innovatively utilizing NLP to automatically detect sectors from budget announcements and rank them according to their predicted performances. This methodological advancement enables the proactive identification of potential market impacts, providing valuable foresight for both investors and policymakers.

\section{Problem Statement}
This study addresses two sequential challenges in computational fiscal analysis:

\begin{enumerate}
    \item \textbf{Multi-Label Sector Classification} \\
    Given a budget transcript segment \( t \in T \) from India's Union Budget corpus (1947--2025), determine the probabilistic association \( P(s_i|t) \) for each sector \( s_i \in S \), where \( S = \{s_1, \ldots, s_{81}\} \) represents formal economic sectors. The task requires overcoming:
    \begin{itemize}
        \item Implicit sector references in policy language (e.g., ``Credit access for handloom industries'' \(\rightarrow\) Banking, Textile sectors)
        \item Domain-specific lexical ambiguity (e.g., ``digital infrastructure'' mapping to both Technology \& Utilities sectors)
    \end{itemize}

    \item \textbf{Performance-Aware Sector Ranking} \\
    For identified sector set \( \hat{S} = \{s_j \mid P(s_j|t) > \tau\} \), develop a model \( f: \hat{S} \rightarrow \mathbb{R}^+ \) that ranks sectors by expected next day post-announcement returns \( r_s \) using text excerpts $t$ related to the sector $s\_j$. Here, $\tau$ represents probabilistic threshold.
\end{enumerate}


\section{Dataset Construction}

\subsection*{Data Collection \& Curation}
\begin{itemize}
    \item \textbf{Sector-Company Mapping}: We systematically collected a list of sectors and their constituent companies from Screener.in. \footnote{\url{https://www.screener.in/explore/} (accessed on 17\textsuperscript{th} March, 2025)}
    \item \textbf{Budget Transcripts}: Aggregated 97 Union Budget documents (1947--2025) from India's Ministry of Finance portal \footnote{\url{https://www.indiabudget.gov.in/bspeech.php} (accessed on 17\textsuperscript{th} March, 2025)}, comprising 1,600+ text excerpts. This also includes the interim budgets.
\end{itemize}

\subsection*{Annotation Pipeline}
\begin{enumerate}
    \item \textbf{Sector Tagging}: For each of the budget transcripts, we prompted DeepSeek \cite{deepseekai2025deepseekr1} to extract texts and corresponding sector(s) as mentioned in \S \ref{sec:budget-prompt}. 
    \item \textbf{Validation}: We manually validated all the outputs.
\end{enumerate}

\subsection*{Market Response Quantification}
For sector $s$ in budget day $d$ of a financial year, performance metric $r_{s,d}$ calculated as:
\[
r_{s,d} = \frac{1}{|C_s|}\sum_{c \in C_s} \frac{P_{c,d+1}^{\text{open}} - P_{c,d}^{\text{open}}}{P_{c,d}^{\text{open}}}
\]
where $C_s$ denotes constituent companies of sectors, with historical data sourced from yahoo finance. \footnote{\url{https://finance.yahoo.com/} (accessed on 17\textsuperscript{th} March, 2025)} $P_{c,d}^{\text{open}}$ denotes the opening price of company c on day d. Finally, we ranked the sectors in decreasing order of their performances. More details about the data is presented in Table \ref{tab:budget-dataset}. Data till the year 2019 was used for training, data spanning 2020 to 2023 was allocated for validation, and 2024 data was reserved for testing.

\begin{table*}[ht]
\centering
\caption{Dataset Statistics}
\label{tab:budget-dataset}
\begin{tabular}{@{}lrrr@{}}
\toprule
\textbf{Metric} & \textbf{Budget Transcripts} & \textbf{Sector Identification} & \textbf{Sector Ranking} \\ \midrule
Total Entries & 97 & 1,671 & 429 \\
Temporal Span & 1947--2025 & 1947--2025 & 1997--2025 \\ \bottomrule
\end{tabular}
\end{table*}

\section{Experiments \& Results}
This study involved two primary experimental components. Firstly, we employed a methodology to identify specific sectors from excerpts of budget transcripts. Secondly, we developed a framework to rank these identified sectors based on their performance, thereby providing a comprehensive analysis of sectoral impacts.

\subsection{Identifying sectors from excerpts of budget transcripts}
The task of identifying sectors from budget excerpts was approached as a multi-class classification problem. We implemented and evaluated several methodologies to address this challenge.

Initially, we employed semantic similarity (STS) based on Nomic embeddings \cite{nussbaum2024nomic} to identify sectors from given text excerpts. To enhance performance, we subsequently fine-tuned these embeddings to optimize the vector space representation, such that sectors relevant to a particular excerpt were positioned closer together, while unrelated sectors were distanced. Additionally, we fine-tuned pre-trained language models, specifically BERT \cite{devlin-etal-2019-bert}, and RoBERTa \cite{Liu2019RoBERTaAR}, for the classification of budget excerpts into appropriate sectors.

The performance metrics for the various models are presented in Table \ref{tab:budget-sector-classify-results}. Our analysis reveals that the STS model with fine-tuned embeddings, and $\tau$ equals to 0.5 demonstrated superior performance in terms of both Macro (M) and Weighted (W) F1 scores. This suggests that the fine-tuned embedding approach effectively captures the nuanced relationships between budget language and sectoral classifications. Conversely, the BERT model exhibited the highest Micro (m) F1 score, indicating its strength in correctly classifying the most frequent sector categories.

\subsection{Ranking Sectors based on their performance}

To rank sectors based on their performance, we developed and evaluated four distinct architectural approaches.

Our initial approach involved transforming sector performance data into a binary classification task, determining whether a given sector would experience an upward or downward movement based on the text excerpts related to it. Using this framework, we fine-tuned three encoder-based (Enc) models: BERT \cite{devlin-etal-2019-bert}, RoBERTa \cite{Liu2019RoBERTaAR}, and DeBERTa \cite{he2020deberta} for classification purposes. The predicted probabilities from these models were then utilized to generate sector rankings.

Building upon this classification approach, we subsequently fine-tuned the same models for regression analysis. This allowed us to predict the actual performance metrics for each sector with greater precision. The sectors were then ranked according to these predicted performance values, providing a more nuanced assessment of relative sectoral strength.

Following our encoder-based approaches, we implemented feature-based models utilizing Nomic embeddings \cite{nussbaum2024nomic} (Emd) extracted from sector-related text excerpts. For binary classification, we trained several machine learning algorithms including logistic regression, random forest, and XGBoost \cite{xgboost}. These models were tasked with predicting whether sectors would experience positive or negative performance.

In parallel, we developed regression models using linear regression, random forest, and XGBoost algorithms to predict the actual performance metrics of each sector. The ranking methodology remained consistent with our previous approaches, wherein sectors were ordered based on their predicted performance values. Additionally, we trained an XGBoost model specifically optimized with a learning-to-rank objective to directly produce sector rankings.

In our final experimental approach, we leveraged state-of-the-art large language models (LLMs) to estimate sector performance based on budget text excerpts. Specifically, we employed three advanced LLMs: Gemma-3 27B \cite{gemma_2025}, DeepSeek V3 \cite{deepseekai2025deepseekv3technicalreport}, and Llama 3.3 70B \cite{touvron2023llamaopenefficientfoundation}. These models were prompted to analyze the sector-relevant text excerpts and estimate the expected performance metrics for each sector. The resulting performance estimates were then utilized to generate sector rankings. More details of the prompts are provided in \S \ref{sec:sector-performance-prompt}.

Table \ref{tab:budget-sector-ranking} presents the comparative performance metrics for these these architectural approaches. Notably, the BERT model trained for classification exhibited superior performance in terms of Normalized Discounted Cumulative Gain (NDCG), suggesting that smaller models are more effective when we have lesser number of instances to train. The performance of the LLMs is comparable to that of the other approaches.

\begin{table}[]
\centering
\caption{Results of Multi-Label Sector Classification}
\label{tab:budget-sector-classify-results}
\resizebox{\columnwidth}{!}{%
\begin{tabular}{lrrr}
\toprule
 & \multicolumn{1}{l}{\textbf{F1 (M)}} & \multicolumn{1}{l}{\textbf{F1 (m)}} & \multicolumn{1}{l}{\textbf{F1 (w)}} \\ 
 \midrule
\textbf{STS (base)}      & 0.159          & 0.176          & 0.345          \\ 
\textbf{STS (fine-tune)} & \textbf{0.291} & 0.478          & \textbf{0.605} \\ 
\textbf{BERT}            & 0.179          & \textbf{0.489} & 0.425          \\ 
\textbf{RoBERTa}         & 0.075          & 0.274          & 0.192          \\ 
\bottomrule
\end{tabular}%
}
\end{table}

\begin{table}[]
\centering
\caption{Sector Ranking Results}
\label{tab:budget-sector-ranking}
\resizebox{\columnwidth}{!}{%
\begin{tabular}{llr}
\toprule
\textbf{Model} & \textbf{Type}    & \multicolumn{1}{l}{\textbf{NDCG}} \\ \midrule
BERT           & Enc Clasifier    & \textbf{0.997}                     \\ 
RoBERTa        & Enc Clasifier    & 0.994                              \\ 
DeBERTa        & Enc Clasifier    & 0.996                              \\ 
BERT           & Enc Regressor    & 0.995                              \\ 
RoBERTa        & Enc Regressor    & 0.995                              \\ 
DeBERTa        & Enc Regressor    & 0.995                              \\ 
Logistic       & Emd + Classifier & 0.996                              \\ 
Random Forest  & Emd + Classifier & 0.996                              \\ 
XG-Boost       & Emd + Classifier & 0.994                              \\ 
Linear         & Emd + Regressor  & 0.995                              \\ 
Random Forest  & Emd + Regressor  & 0.996                              \\ 
XG-Boost       & Emd + Regressor  & 0.994                              \\ 
XG-Boost       & Learning to Rank & 0.994                              \\ 
Gemma-3 27B      & Zero Shot & 0.994                              \\ 
DeepSeek V3       & Zero Shot & 0.993                              \\ 
Llama 3.3 70B       & Zero Shot & 0.994                              \\ \bottomrule
\end{tabular}%
}
\end{table}

\section{Conclusion}
This study presents a comprehensive framework for the detection and performance-based ranking of sectors from Indian Union Budget transcripts. Our findings demonstrate that fine-tuned Nomic-based embeddings provide superior performance in identifying sectors from textual excerpts, capturing the nuanced relationships between budget language and sectoral classifications. Concurrently, the BERT-based model fine-tuned for classification emerged as the most effective approach for ranking sectors based on predicted performance, surpassing other methodologies.

The framework developed in this research offers valuable insights for investors, and financial analysts seeking to understand the immediate market implications of budget announcements. By automating the process of sector identification and performance prediction, our approach enables more timely and informed decision-making.

Future research directions include extending this framework to recommend specific stocks within the identified sectors, potentially offering more granular investment guidance. Additionally, developing capabilities to capture real-time price movements following budget announcements would enhance the practical applicability of this work. 

\bibliography{custom}

\section*{Limitations}

\label{sec:limitations}

Despite our methodological contributions, several limitations warrant acknowledgment. 

First, our annotation approach emphasized precision over recall in sector identification. The DeepSeek language model may have overlooked subtler budget-sector relationships, particularly when policy implications were implicit rather than explicit. Our validation protocol—focusing exclusively on LLM-detected relationships—potentially reinforces this detection bias, creating systematic blind spots in the dataset. Consequently, fiscal impacts on certain sectors may be underrepresented in our analysis.

Second, temporal coverage presents significant constraints. Market performance data availability beginning only from 1997 excluded 50 years of budget documents (1947-1996) from complete analysis. This limitation is particularly significant when analyzing long-term policy impacts and historical shifts in sector prioritization. Additionally, inconsistent market data across sectors forced the exclusion of certain sector-period combinations, introducing potential selection bias. These gaps disproportionately affected newly formalized sectors and those with limited public listings.

Third, our performance metric isolates budget effects without controlling for confounding variables. Macroeconomic factors (monetary policy adjustments, global market movements), sector-specific events (regulatory changes, technological disruptions), and concurrent corporate announcements likely influence post-budget market movements. The absence of a comprehensive control framework limits causal interpretations of budget-performance relationships.

Future research should address these limitations through multi-source validation, synthetic data generation for pre-1997 periods, and development of counterfactual models that control for non-budgetary market influences.

\appendix

\section{Appendix}
\label{sec:budget-appendix}
\subsection{Industries}
List of industries are as follows: [`Aerospace \& Defence' , `Agro Chemicals' , `Air Transport Service' , `Alcoholic Beverages' , `Auto Ancillaries' , `Automobile' , `Banks' , `Bearings' , `Cables' , `Capital Goods - Electrical Equipment' , `Capital Goods-Non Electrical Equipment' , `Castings ,  Forgings \& Fastners' , `Cement' , `Cement - Products' , `Ceramic Products' , `Chemicals' , `Computer Education' , `Construction' , `Consumer Durables' , `Credit Rating Agencies' , `Crude Oil \& Natural Gas' , `Diamond ,  Gems and Jewellery' , `Diversified' , `Dry cells' , `E-Commerce/App based Aggregator' , `Edible Oil' , `Education' , `Electronics' , `Engineering' , `Entertainment' , `Ferro Alloys' , `Fertilizers' , `Finance' , `Financial Services' , `FMCG' , `Gas Distribution' , `Glass \& Glass Products' , `Healthcare' , `Hotels \& Restaurants' , `Infrastructure Developers \& Operators' , `Infrastructure Investment Trusts' , `Insurance' , `IT - Hardware' , `IT - Software' , `Leather' , `Logistics' , `Marine Port \& Services' , `Media - Print/Television/Radio' , `Mining \& Mineral products' , `Miscellaneous' , `Non Ferrous Metals' , `Oil Drill/Allied' , `Packaging' , `Paints/Varnish' , `Paper' , `Petrochemicals' , `Pharmaceuticals' , `Plantation \& Plantation Products' , `Plastic products' , `Plywood Boards/Laminates' , `Power Generation \& Distribution' , `Power Infrastructure' , `Printing \& Stationery' , `Quick Service Restaurant' , `Railways' , `Readymade Garments/ Apparells' , `Real Estate Investment Trusts' , `Realty' , `Refineries' , `Refractories' , `Retail' , `Ship Building' , `Shipping' , `Steel' , `Stock/ Commodity Brokers' , `Sugar' , `Telecomm Equipment \& Infra Services' , `Telecomm-Service' , `Textiles' , `Tobacco Products' , `Trading' , `Tyres']
\subsection{Prompts}
\subsubsection{Text Extraction and Sector Identification}
\label{sec:budget-prompt}
You are provided with the budget of India below. From this budget only pick up text segments relevant to the given list of industries.
List of industries: <list of industries>
Your output should be a json file having 2 keys: `text\_segment' and `industry'. The value corresponding to `text\_segment' would be the extract text segment extracted from the budget. The value of `industry' should be the corresponding list of industries from the given list that the text segment is related to. Return only the segments having any relation with the given list of industries. One text segment can be related to multiple industries.

Text context from Budget: <Budget Transcript of a given year>

\subsubsection{Sectorwise Performance Prediction}
\label{sec:sector-performance-prompt}
You are a financial expert with extensive experience of analysing Indian Budgets. Given a sector and an excerpts related to the sector from a budget speech, estimate the performance of the sector. You output should be just a real number between -1 to 1. Don't reply anything else. Sector: <name of sector>, Excerpt: <text excerpts related to the given sector>

\subsection{Reproducibility}
The codes and the datasets can be accessed from \url{https://huggingface.co/datasets/sohomghosh/BASIR_Budget_Assisted_Sectoral_Impact_Ranking/tree/main/}

\end{document}